\pdfoutput=1

\documentclass[11pt]{article}

\usepackage[final]{acl}

\usepackage{times}
\usepackage{latexsym}

\usepackage[T1]{fontenc}

\usepackage[utf8]{inputenc}

\usepackage{microtype}

\usepackage{inconsolata}
\usepackage{lipsum}  

\usepackage{graphicx}
\usepackage{algorithm}
\usepackage{algpseudocode}
\usepackage{amsmath}

\usepackage{xcolor}

\title{Multimodal Contextualized Semantic Parsing from Speech}

\author{Jordan Voas \and Raymond Mooney \and David Harwath\\
  \texttt{jvoas@utexas.edu} \and
  \texttt{mooney@utexas.edu} \and
  \texttt{harwath@utexas.edu} \\
  The University of Texas at Austin \\}

\begin{document}
\maketitle
\begin{abstract}
We introduce Semantic Parsing in Contextual Environments (SPICE), a task designed to enhance artificial agents' contextual awareness by integrating multimodal inputs with prior contexts. SPICE goes beyond traditional semantic parsing by offering a structured, interpretable framework for dynamically updating an agent's knowledge with new information, mirroring the complexity of human communication. We develop the VG-SPICE dataset, crafted to challenge agents with visual scene graph construction from spoken conversational exchanges, highlighting speech and visual data integration. We also present the Audio-Vision Dialogue Scene Parser (AViD-SP) developed for use on VG-SPICE. These innovations aim to improve multimodal information processing and integration. Both the VG-SPICE dataset and the AViD-SP model are publicly available. \footnote{https://github.com/jvoas655/VG-SPICE} \footnote{https://github.com/jvoas655/AViD-SP}
\end{abstract}

\section{Introduction}
\begin{figure*}
    \centering
    \includegraphics[width=0.9\linewidth]{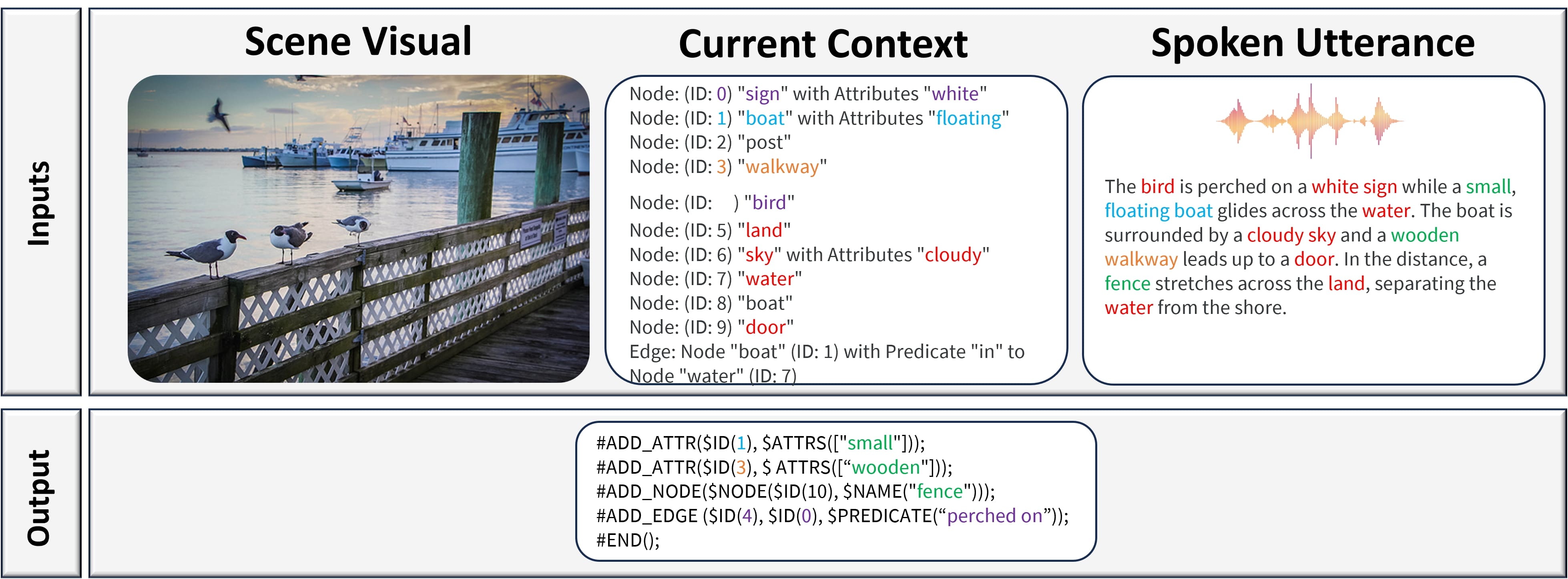}
    \caption{Example of VG-SPICE inputs as well as a plausible output to produce the correct next state context. New information that the agent is expected to add to the context is shown in green while already known information is noted in red. Grounding entities that have new information being added to them are noted in blue and orange. The current context is shown as a textually prompted representation of the actual knowledge graph (discussed in Section \ref{sec: Contextual State Representation}). }
    \label{fig:vg-spice-examples}
\end{figure*}

Imagine you are taking a guided tour of an art museum. During the tour as you visit each piece of art, your guide describes not only the artworks themselves but also the history and unique features of the galleries and building itself. Through this dialog, you are able to construct a mental map of the museum, whose entities and their relationships with one another are grounded to their real-world counterparts in the museum. We engage in this type of iterative construction of grounded knowledge through dialog every day, such as when teaching a friend how to change the oil in their car or going over a set of X-rays with our dentist. As intelligent agents continue to become more ubiquitous and integrated into our lives, it is increasingly important to develop these same sorts of capabilities in them.

Toward this goal, this work introduces Semantic Parsing in Contextual Environments (SPICE), a task designed to capture the process of iterative knowledge construction through grounded language. It emphasizes the continuous need to update contextual states based on prior knowledge and new information. SPICE requires agents to maintain their contextual state within a structured, dense information framework that is scalable and interpretable, facilitating inspection by users or integration with downstream system components. SPICE accomplishes this by formulating updates as Formal Semantic Parsing, with the formal language defining the allowable solution space of the constructed context.

Because the SPICE task is designed to model real-world and embodied applications, such as teaching a mobile robot about an environment or assisting a doctor with medical image annotations, there are crucial differences between SPICE and traditional text-based semantic parsing. First, SPICE considers parsing language within a grounded, multimodal context. The language in cases like these may have ambiguities that can only be resolved by taking into account multimodal contextual information, such as from vision.

Furthermore, SPICE supports linguistic input that comes in the form of both speech and text. In real-world embodied interactions, language is predominantly spoken, not written. While modern automatic speech recognition (ASR) technology is highly accurate, it is still sensitive to environmental noise and reverberation, and representing the input language as both a waveform as well as a noisy ASR transcript can improve robustness. While we do not consider it here, the SPICE framework also supports paralinguistic input such as facial expressions, eye gaze, and hand gestures.

We present a novel dataset, VG-SPICE, derived from the Visual Genome \citep{krishna2016visual}, an existing dataset comprised of annotated visual scene graphs representing constituent entities and relational prepositions, enhanced with additional processing and synthetic augmentation to form a foundational representation for SPICE tasks. VG-SPICE simulates the conversational construction of visual scene graphs, wherein a knowledge graph representation of the entities and relationships contained within an image must be collected from the visual inputs and audio dialogue. This dataset, along with an initial model trained for VG-SPICE, sets the baseline for future efforts. Figure \ref{fig:vg-spice-examples} shows an example of a typical VG-SPICE sample. The figure shows how potential semantic parses can be extracted from the visual scene and spoken utterance conditioned on what information is already known about the scene. 

The remainder of this paper is structured as follows: It begins with a detailed analysis of the SPICE task, introduces the VG-SPICE dataset, and presents our AViD-SP model. It then delves into experimental results, showcasing the model’s ability to process and interpret context consistent with the SPICE framework. Finally we outline the implications and directions for future research. The main contributions include:

\begin{itemize}
    \item A definition of the Semantic Parsing in Contextual Environments (SPICE) task, highlighting its challenges, scope, and significance in enhancing human-AI communication.
    \item The creation of a large, machine-generated SPICE dataset, VG-SPICE, leveraging existing machine learning models and the Visual Genome dataset, to motivate SPICE research.
    \item An initial baseline model, Audio-Vision Dialogue Scene Parser (AViD-SP), for VG-SPICE that integrates Language Models with Audio/Visual feature extractors, establishing a research benchmark for SPICE. As a component of AViD-SP, we also introduce a novel pretrained encoder adaption and multimodal fusion method, the Grouped Multimodal Attention Down Sampler (GMADS) to motivate the exploration of additional multimodal adaptation methods. 
\end{itemize}

\section{Related Work}
The SPICE task intersects with research in dialogue systems and semantic parsing. While previous efforts in these areas have addressed some elements of SPICE, none have fully encapsulated the comprehensive requirements of the SPICE task.

\subsection{Dialogue Systems and Multimodality}
Dialogue systems share similarities with SPICE tasks, particularly in their aim to emulate human conversational skills, including referencing prior conversational context. However, SPICE differentiates itself by necessitating multimodal interactions, the utilization of structured and interpretable knowledge representations, and the capability for dynamic knowledge updates during conversations, setting it apart from conventional dialogue models.

Recent advancements in dialogue systems, particularly through large language models (LLMs) \citep{wei2022finetuned, chowdhery2022palm, ouyang2022training, jiang2023mistral, touvron2023llama, touvron2023llama2}, have enhanced the ability to manage complex, multi-turn conversations. This is largely thanks to the employment of extensive context windows \citep{dao2023flashattention2}, improving language comprehension and generation for more coherent and contextually appropriate exchanges. Nevertheless, LLMs' reliance on broad textual contexts can compromise efficiency and interpretability in many applications. Not only must all prior inputs be reprocessed for future updates but the uncompressed format prevents easy end-user inspection of the information the model is tracking for future interactions. 

Advances in multimodal dialogue systems, incorporating text, image, and audio inputs \citep{liu2023visual, zhu2023minigpt4, dai2023instructblip, zhang2023videollama, maaz2023videochatgpt}, edge closer to SPICE's vision of multimodal communication. Yet, these systems cannot often distill accumulated knowledge into concise, understandable formats, instead still relying on raw dialogue histories or opaque embeddings for prior context.

While some systems are beginning to interact with and update external knowledge bases, these interactions tend to be unidirectional \citep{cheng-etal-2022-hitab, wu-etal-2021-better} or involve knowledge storage as extensive, barely processed texts \citep{zhong2023memorybank, wang2023recursively}. Dialogue State Tracking (DST) \citep{balaraman-etal-2021-recent} shares similarities with SPICE in that agents use and update their knowledge bases during dialogues. However, most DST efforts are unimodal, with limited exploration of multimodal inputs \citep{kottur-etal-2021-simmc}. Moreover, existing datasets and models for DST do not align with the SPICE framework, as they often rely on regenerating the knowledge base with each dialogue step from all historical dialogue inputs without offering a structured representation of the prior context. SPICE, conversely, envisions sequential updates based on and directly applied to prior context, a feature not yet explored in DST. Further, we are unaware of any DST work that has attempted to utilize spoken audio.

\subsection{Semantic Parsing}
Semantic Parsing involves translating natural language into a structured, symbolic-meaning representation. Traditional semantic parsing research focuses on processing individual, short-span inputs to produce their semantic representations \citep{kamath2019survey}. Some studies have explored semantic parsing in dialogues or with contextual inputs, known as Semantic Parsing in Context (SPiC) or Context Dependent Semantic Parsing (CDSP) \citep{li-etal-2020-context}. However, most CDSP research has been aimed at database applications, where the context is a static schema \citep{yu-etal-2019-cosql}. While these tasks leverage context for query execution, they do not involve dynamic schema updates, instead maintaining a static context between interactions. Outside these applications, CDSP is mainly applied in DST \citep{ye2021slot, cheng-etal-2020-conversational, moradshahi-etal-2023-contextual, heck-etal-2020-trippy}, which we have previously differentiated from SPICE.

Furthermore, semantic parsing has traditionally been limited to textual inputs and unimodal applications. It has been extended to visual modalities, notably in automated Scene Graph Generation (SGG) tasks \citep{Zhang2023LearningTG, abdelsalam-etal-2022-visual, zareian2020weakly}. Although there has been exploration into using spoken audio for semantic parsing \citep{tomasello2022stop, coucke2018snips, lugosch2019speech, sen-groves-2021-semantic}, these efforts have been constrained by focusing on simple intent and slot prediction tasks, and have not incorporated contextual updates or complex semantic outputs.

As such, we believe SPICE to be considerably distinct from any works that have come previously. While individual components of SPICE's framework have been studied, such as semantic parsing from audio, context, or multimodal inputs, no work has utilized all of these at once. Additionally, SPICE goes beyond most semantic parsing and dialogue works, even those operating on some form of knowledge representation, by tasking the agent to produce continual updates to said knowledge graph and to maintain them in an interpretable format.

\begin{table*}
    \centering
    \begin{tabular}{lcccc}
        \hline
        \textbf{Dataset} & \textbf{\#Scenes} & \textbf{\#Nodes} & \textbf{\#Predicates} & \textbf{Avg. Size}\\
        \hline
         Visual Genome \citep{krishna2016visual} & 108077 & 76,340 & - & - \\
         VG80K \citep{zhang2019largescale} & 104832 & 53304 & 29086 & 19.02 \\
         VG150 \citep{xu2017scene} & 105414 & 150 & 50 & 6.98 \\
         \hline
         Ours & 22346 & 2032 & 282 & 19.64 \\
        \hline
    \end{tabular}
    \caption{\label{tab:vg-clean}
        Comparison of our Visual Genome curation statistics to other works. Further details are in Section \ref{sec: Visual Genome Preprocessing}.
    }
\end{table*}

\section{Task Definition}

Semantic Parsing in Contextual Environments (SPICE) is defined as follows. Consider a model agent, denoted as $a$, designed to maintain and update a world state across interaction timesteps. Let $C_{i}$ represent this world state during the $i^{th}$ turn. For interpretability and downstream use $C_{i}$ is represented as a formal knowledge graph \citep{CHEN2020112948}. This state represents the accumulated context from prior interactions. Initially, $C_{i}$ can be set to a default or empty state.

During each interaction turn, the agent encounters a set of new inputs, referred to as information inputs $F_{i}^{m}$, with $m$ indicating the diversity of modalities the agent is processing. The agent's goal is to construct a formal semantic parse, $P_{i} = a(F_{i}^{m}, C_{i})$. This parse is formulated by integrating the prior context $C_{i}$ with the new information inputs $F_{i}^{m}$. With the aid of an execution function $e$, this results in an updated context $C_{i+1}$ = $e(P_{i}, C_{i})$. 

This newly formed context $C_{i+1}$ should represent all task essential information, both from previous context $C_{i}$ and the most recent interaction round, for future rounds. $C_{i+1}$ is expected to align with a reference context, denoted as $\hat{C}_{i+1}$, which represents the ideal post-interaction state.

\section{Dataset}

This section introduces VG-SPICE, a novel dataset for SPICE tasks, providing a structured benchmark for model training and evaluation. To our knowledge, VG-SPICE is the first of its kind and is derived from the Visual Genome dataset \citep{krishna2016visual} to simulate a ``tour guide'' providing sequential descriptions of aspects of the environment. In these scenarios, the tour guide describes a visual scene with sequential utterances, each introducing new elements to the scene. These descriptions, combined with a pre-established world state of the scene, mimic the accumulation of world state information through successive interactions.

VG-SPICE utilizes the Visual Genome's 108k images with human-annotated scene graphs for entity identification via bounding boxes, originally detected using an object identification model. The graphs include named nodes, optional attributes, and directed edges for relational predicates.

The dataset is constructed by extracting sub-graphs from scene graphs as the initial context, $C_{i}$, sampled from empty to nearly complete. These are then augmented by reintegrating a portion of the omitted graph to form the updated context, $C_{i+1}$. Before extracting our samples, the Visual Genome data underwent preprocessing to enhance dataset quality (Section \ref{sec: Visual Genome Preprocessing} and summary results shown in Table \ref{tab:vg-clean}). The dataset allows flexible model implementation with semantic parses ($P_{i}$) and parsing functions ($e$) not predefined, allowing flexibility in modeling implementation. Our model's semantic parse format is discussed in Section \ref{sec: Formal Language Definition}.

For each context pair ($C_{i}$, $C_{i+1}$), features from $C_{i}$ and modified features for $C_{i+1}$ are structured into natural language prompts. These prompts are processed by the Llama 2 70B LLM \citep{touvron2023llama} to generate plausible sentences that describe the difference between $C_{i}$ and $C_{i+1}$. We then synthesize spoken versions of these sentences via the Tortoise-TTS-V2 \citep{Betker_TorToiSe_text-to-speech_2022} text-to-speech (TTS) synthesis system. We configure the TTS model to randomly sample speaker characteristics from its pretrained latent space, and use the built-in ``high\_quality'' setup for other generation settings. Before TTS conversion filtering is performed on the textual utterances to remove common recurrent terms indicative of new information (eg., "there now is a" versus "there is a"). The audio recordings and visual images are the multimodal inputs $F_{i}^{m}$ of VG-SPICE, emphasizing spoken audio for practicality in real-world applications and necessitating addressing the challenges of semantic parsing from audio such as speaker diversity and noise robustness. The presence of both textual and spoken audio representations for the update utterances allows VG-SPICE to be utilized for semantic parsing evaluations in either modality. 

VG-SPICE includes over 131k SPICE update samples from 20k unique scenes, with $2.5\%$ allocated to each of the validation and test sets, ensuring distinct scenes across splits. We perform noise augmentation on the input speech using the CHiME5 dataset \citep{barker2018fifth} to simulate realistic noise conditions, with performance evaluated at various Signal to Noise Ratios (SNR). VG-SPICE samples and summary statistics are presented in Figure \ref{fig:vg-spice-examples} and Table \ref{tab:vg-spice-stats}, respectively.

\begin{table}
    \centering
    \begin{tabular}{lc}
        \hline
        \textbf{Statistic} & \textbf{Value}\\
        \hline
         \# Samples & 131362 \\
         \# Unique Scenes & 22346 \\
         Hours of Audio & 10.56 \\
         Avg. Words per Utterance & 71.83  \\
         Avg. Nodes Added & 1.27 \\
         Avg. Attributes Added & 0.93 \\
         Avg. Edges Added & 0.60 \\
        \hline
    \end{tabular}
    \caption{\label{tab:vg-spice-stats}
        Summary statistics for our VG-SPICE dataset.
    }
\end{table}

\subsection{Challenge Subset}

In addition to the standard test set, we augment VG-SPICE with an additional Challenge Subset, VG-SPICE-C. Although this subset is small, spanning only 50 individual visual scenes, it provides distinct capabilities not present in the primary VG-SPICE test dataset, as detailed below. 

\textbf{Broad Visual Representation}: To sample the Challenge Subset, we used a representation-based process to promote diverse image types. We obtained the CLIP\footnote{openai/clip-vit-base-patch32 from Huggingface} representations for each image in the original VG-SPICE test split. Using KMeans clustering, the dataset was partitioned into 50 distinct groupings of visual representations, with a single sample taken from each cluster.

\textbf{Manual Scene Graph Quality Enhancements}: Despite automated generation processes in VG-SPICE aiming to improve scene graph quality, persistent issues remain. To ensure a clean and reliable testing subset, manual scene graph improvements were made to ensure the final scene graph for each image was accurate. This involved removing incorrect, low-quality, or duplicate scene features and enhancing the scene graphs to achieve far greater density than originally present in VG-SPICE or Visual Genome, particularly for Edges and Attributes.

\textbf{Coherent Iterative Updates}: To improve sample diversity, VG-SPICE was generated in an iteratively incoherent fashion, meaning samples for a single update cannot be used to coherently evaluate end-to-end SPICE evaluations. For the Challenge Subset, we manually annotated each of the 50 sampled scenes with five individual utterances, each adding novel information while referring to previously mentioned details. These utterances are of greater diversity and quality (due to manual annotation rather than LLM production) and can be used sequentially to evaluate scene graph generation errors over multiple interaction rounds.

\textbf{OOD and Real Speech}: To enhance the evaluative capabilities of the Challenge Set, we provide speech samples for the utterances from two sources: Tortoise-TTS as used for the remainder of VG-SPICE (with three random voice samples per utterance) as well as manual recordings of the spoken utterances by a individual human annotator.

This Challenge Subset offers a rigorous evaluation framework for models, promoting advancements in handling diverse visual representations, maintaining high-quality scene graphs, performing coherent iterative updates, and managing out-of-domain and real-world speech scenarios.

\section{AViD-SP Model} \label{AViD-SP Model}

\begin{figure}[t]
\centering
\includegraphics[width=1.0\linewidth]{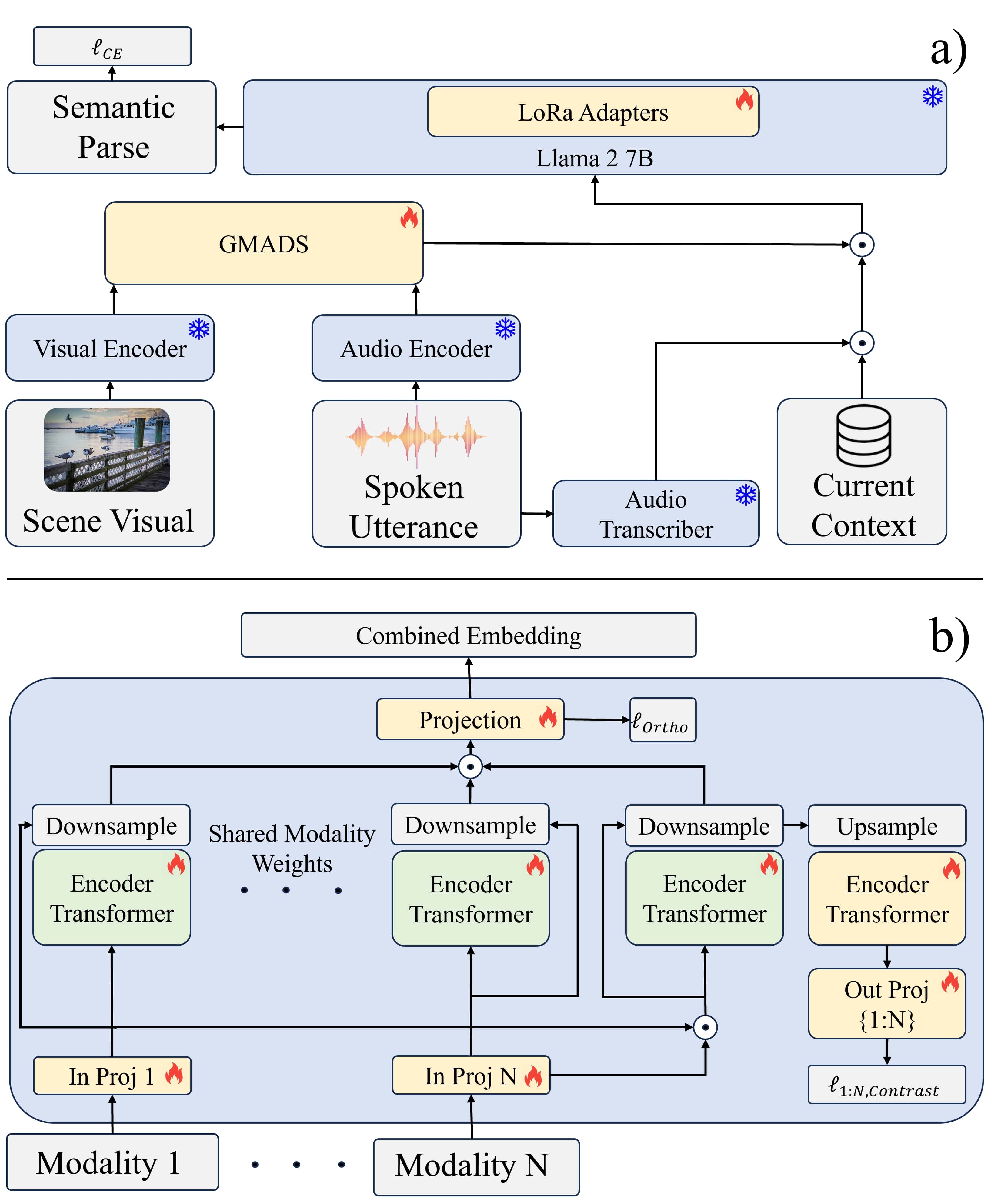}
\caption{a) The architecture of the AViD-SP model for VG-SPICE, integrating pretrained encoders and large language models (LLMs) with LoRa adapters and feature fusion modules. Trained and frozen segments of the model are denoted by fire and snowflake icons, respectively. b) Our novel Grouped Modality Attention Down Sampler module, enabling integrated cross-modality fusion and downsampling. Green modules share weights. For downsampling, we utilize meanpooling, and for upsampling we linearly interpolate the embeddings. }
\label{fig:vg-spice-model2}
\end{figure}

To address the challenges of VG-SPICE, our approach utilizes a range of pretrained models, specifically fine-tuned to enhance SPICE-focused semantic parsing capabilities. Figure \ref{fig:vg-spice-model2} illustrates our model architecture, termed Audio-Vision Dialogue Scene Parser (AViD-SP). At the core of our framework lies the pretrained Llama 2 7B model \citep{touvron2023llama2}. Despite deploying its smallest variant, the extensive pretraining endows our model with robust functional abilities, particularly beneficial for processing the diverse semantic parses inherent to VG-SPICE. However, Llama 2, trained on textual data, lacks inherent support for the multimodal inputs typical in VG-SPICE.

To accommodate diverse inputs, we extend techniques from prior studies \citep{rubenstein2023audiopalm, gong2023listen, lin2023sphinx} by projecting embeddings from pretrained modality-specific feature extractors. This approach has been proven to enable text-based LLMs to process information across various modalities. Directly integrating these projected embeddings into the LLM's context window, however, introduces significant computational overhead due to their typically extensive context lengths. While previous research often employed pooling methods \citep{gong2023listen} to condense embeddings by modality, this strategy incompletely addresses the challenges of merging varied modality embeddings for LLM use. For instance, audio embeddings offer finer temporal granularity than textual embeddings, and the reverse is often true for vision embeddings, complicating the adjustment of downsampling factors. Moreover, even with optimized downsampling, pooled embeddings must preserve their original sequential order and are restricted to information from solely the pooled segments. Many applications could benefit from capabilities to establish downsampled features encompassing both local and global contexts and to rearrange these features to an extent.

To surmount these challenges, we introduce a novel Grouped Modality Attention Down Sampler (GMADS) module. This module initially projects embeddings from non-textual modalities into a unified, fixed-dimensional space. We form a set of modality groupings, one for each input modality (audio and visual with VG-SPICE), and a cross-modality grouping derived from concatenating all modality embeddings, each prefixed with a modality-specific token. A series of self-attention layers processes each embedding sequence and downsamples the outputs by a factor of $S$ through mean pooling. These values are then concatenated with the mean-pooled pre-self-attention embeddings along the embedding dimension, akin to a skip connection. A final projection adjusts the outputs to match the dimensionality of the Llama 2 7B decoder, and all embedding sequences are concatenated. This process yields an embedding output that is effectively downsampled by a factor of $S/2$. All weights in the GMADS module are shared across the groups, substantially reducing the parameter count. Additionally, we employ a self-supervised representation learning objective on the embeddings from the downsampled cross-modality group outputs by upsampling them to their original size and then processing them through a secondary set of self-attention layers. The reconstructed cross-modality embeddings are then segmented by modality, with per-modality projections striving to restore them to their original input size. We apply a contrastive reconstruction loss objective as outlined in Eq. \ref{eq:contrastive-loss}, using the corresponding ground truth embedding as an anchor and all other embeddings in the batch as contrastive samples. 
\begin{multline}
    \label{eq:contrastive-loss} \ell_{n, Contrast} = \scriptstyle{\sum_{j=1}^{B*K} \log \frac{\exp(sim(z_i, z_j)/\tau)}{\sum_{k=1}^{B*K} [k \neq i] \exp(sim(z_i, z_k)/\tau)}}
\end{multline}
In this equation $z_{i}$ denotes the reconstructed input embedding, $K$ represents the length of each sequence, $B$ denotes the batch size, and $\tau$ is a tunable temperature hyperparameter.

We also observed that non-textual modality inputs tended to collapse when combined with simpler textual inputs, such as prior context or ASR transcripts. To counter this, we include an additional orthogonality loss, designed to encourage maximal dissimilarity among aligned embeddings in each batch sequence. This methodology is similar to previous efforts to promote distinct class embeddings \cite{ranasinghe2021orthogonal}, but in our case, we treat each embedding as a distinct class sample. However, given the nature of these embedding sequences, some level of similarity is expected, and entirely dissimilar values (cosine similarity less than zero) are not feasible. Thus, we modify Eq. \ref{eq:ortho-loss} to include a slight margin allowing for minimal similarity. Below, $e_i$ represents a single GMADS output embedding (pre-output projection) within a batch of $B$ sequences, each of length $K$.

\begin{multline}
    \label{eq:ortho-loss} \ell_{Ortho} =  {\scriptstyle \frac{2\sum_{i=1}^{B*K-1} \sum_{j=i+1}^{B*K} max(\frac{e_{i} * e_{j}}{\|e_{i}\| * \|e_{j}\|} - h, 0))}{B*K * (B*K-1)}}
\end{multline}

The GMADS module attempts to provide several advantages over the direct use of raw modality embeddings with the LLM decoder or mean pooling. Firstly, GMADS operates at reduced dimensional scales compared to the pretrained LLM, which significantly lowers memory requirements, requiring the much larger decoder to process shorter (reduced to only $2/S$ the size) input sequences. Moreover, the modality inputs do not necessitate autoregressive generation alongside these inputs, further conserving cost. Secondly, GMADS empowers the model to selectively learn its downsampling process, including choices on whether to focus locally or integrate global features, allowing some degree of information restructuring. The incorporation of cross-modality encoding enables parts of the downsampled embeddings to capture essential information across modalities while maintaining individual modality components in the outputs ensuring that some portion of the output embeddings is conditioned on each modality, requiring the attention mechanisms to remain sensitive to all modalities.

For feature extraction, we utilize the visual encoder from DINOv2 \citep{oquab2024dinov2} for visual inputs and the encoder from Whisper-Large V3 \citep{radford2022robust} for audio. We retain only the necessary encoder portions of these pretrained models. In alignment with successful semantic parsing efforts from speech \citep{Arora2023IntegratingPA}, we perform ASR transcription on the audio, appending these textual embeddings to the prior context embeddings. ASR transcriptions are generated using the Whisper-medium.en model. To enable scalable fine-tuning, we integrate LoRa adaptation layers into Llama 2 7B and freeze all feature extractors.

\subsection{Training Routine}

We train AViD-SP using cross-entropy loss (Eq. \ref{eq:ce-loss}) between the predicted and reference Formal Semantic Parses, alongside the objectives in Eq. \ref{eq:contrastive-loss} and \ref{eq:ortho-loss}. Our comprehensive loss function is outlined below in Eq. \ref{eq:full-loss}, where $p_{i,k}$ denotes the softmax prediction for each of the $k$ tokens in $P_{i}$, and $t_{i,k}$ represents the corresponding ground-truth token label.

\begin{equation}
    \label{eq:ce-loss} \ell_{CE} = -\sum_{k=1}^{n} t_{i, k} \log(p_{i, k})
\end{equation} 
\begin{equation}
    \label{eq:full-loss} L = \alpha \ell_{CE}+ \beta \ell_{Ortho} + \frac{\gamma}{N} \sum_{n=1}^{N}\ell_{n,Contrast}
\end{equation} 

AViD-SP employs a three-layer self-attention transformer as the primary encoder transformer, each layer having an embedding dimensionality of 1024 and 8 attention heads. The secondary encoder transformer, used for the upsampled reconstruction training objective, is of the same configuration. The GMADS module employs a downsampling factor, $S$, of 16. Additionally, we enhance the key, query, and value layers of the Llama 2 7B model with Low-Rank Adaptation (LoRa) layers. No hyperparameter optimization was conducted.

We train AViD-SP by incorporating randomly sampled CHiME5 noise to simulate audio corruption, adding this noise at various Signal-to-Noise Ratios (SNR) of 0, 2, 5, 10, or 20dB. Further details on training and inference hyperparameters are discussed in Section \ref{sec: Training and Inference Hyperparameters}. To ensure robustness to various input feature combinations, we implement random input dropout with a probability of 30\%. In these instances, we randomly omit one of the input modalities, either audio embeddings, visual embeddings, or audio transcriptions. We do not omit the prior context, as we found the task too difficult to learn under such conditions since it requires both the already known information as well as their current assigned labels under our semantic parsing framework. AViD-SP is trained in a two-stage pipeline, with the initial stage acting as pretraining without the ASR transcriptions to allow the GMADS module to reach a semi-trained state for enhanced efficiency. Subsequently, we continue fine-tuning the model with ASR transcriptions until convergence. Our initial pretraining lasts one full epochs, followed by the fine-tuning stage.

\subsection{Evaluation Metrics}

We use several metrics to measure how closely the generated semantic parse aligns with the ground truth and how accurately the scene graph context updates match the reference. Unlike conventional semantic parsing assessments \citep{tomasello2022stop}, we omit exact-match metrics due to their unsuitability for our problem, which allows for permutation invariance in the formal-language output (see Section \ref{sec: Formal Language Definition}). This permits the parser to generate scene-graph updates in any order and assign node IDs freely, as long as the resulting scene graph is isomorphic to the reference.

For each below metric, we examine hard ("H") and soft ("S") variants. The hard variant penalizes missing and unnecessary information, while the soft variant only penalizes omissions. This approach accounts for the Visual Genome dataset's sparsity and the possibility of LLMs generating extraneous yet potentially valid content. For example, an LLM might enhance a "blue table" to a "vibrant blue table," making "vibrant" an acceptable attribute. Our analysis shows such inclusions are common in the VG-SPICE dataset, leading us to focus on the soft metric and qualitatively show in Section \ref{sec:results} how updated utterances accommodate these extraneous additions. We include results for GED in the supplement Table \ref{tab:vg-spice-ged-results}.

\paragraph{Graph Edit Distance (GED):} GED calculates the normalized cost to transform the predicted context to the reference one, considering only perfectly semantically equivalent Nodes, Attributes, and Edges. Missing or extra Nodes or Edges increase the error by one, while incorrect Attributes have a smaller penalty of 0.25. GED is not normalized and should be interpreted as the magnitude of incorrect features compared with the reference solution and not as a recall or precision metric. GED is particularly reliant on exact matches, so minor discrepancies (like "snow board" vs. "snowboard") can incur significant penalties, with misalignments doubly penalized in the hard variant. 

\paragraph{Representation Edit Distance (RED):} RED addresses the limitations of GED by employing a ``softer'' semantic similarity to evaluate entity pairings. Using a transformer model for sentence semantic similarity\footnote{ The ``en\_stsb\_roberta\_base'' model from https://github.com/MartinoMensio/spacy-sentence-bert}, RED groups Nodes and their Attributes into descriptive phrases (for example, a "table" Node with "vibrant" and "blue" Attributes becomes "vibrant blue table") and assesses the dissimilarity between potential pairings, using an exhaustic search for optimal pairings of Nodes and Edges. Unmatched Nodes and Edges are considered entirely dissimilar. Since unmodified graph portions from the prior context are pre-matched and excluded from the exhaustive search, the computation of the pairings remains manageable. RED is normalized by the representation edit distance needed to transform the prior context into the reference context, and so numerically can be interpreted as the percentage of missing and/or extra information relative to the reference context. 

\subsection{Baselines and Evaluation}

To thoroughly evaluate our AViD-SP model, we conducted a series of ablation studies to explore the impact of various input modality combinations. Given that AViD-SP was trained under diverse noise conditions, its performance was tested across noise levels of 0, 2, and 20 dB using the CHiME5 dataset. We assessed the model's capability to resolve ambiguities in audio input by introducing tests with and without visual modality, and by evaluating the model with incorrectly matched images in the GMADS module. Additionally, we explored potential enhancements in ASR performance by incorporating ground truth ASR transcriptions in our evaluations. To ablate the effects our GMADS module has on performance we compare against a version of AViD-SP trained using traditional meanpooling after a per modality projection layer to downsample the audio and visual input embeddings, with all hyperparameter and training methods matched between the two except the meanpooling baseline only utilizing the cross entropy component of the full training objective. 

We also extended our evaluations to the VG-SPICE-C Subset. Here, we analyze the subset through a single-step evaluation approach, with ground truth prior context provided and metrics measured after each individual SPICE update.  


\begin{table*}
    \small
    \centering
    \begin{tabular}{lr | cccc | cccc}
        \hline
         \multicolumn{2}{l}{Model Type} & \multicolumn{4}{| c |}{H-RED$\downarrow$} & \multicolumn{4}{| c }{S-RED$\downarrow$} \\
         \multicolumn{2}{l |}{} & 0dB & 2dB & 20dB & Gold* & 0dB & 2dB & 20dB & Gold* \\
         \hline
         AViD-SP + GMADS & Base & 1.618 & 1.517 & 1.412 & 1.272 & 0.402 & 0.383 & 0.3765 & 0.348 \\
          & w/o Image & 1.611 & 1.527 & 1.430 & 1.33 & 0.407 & 0.393 & 0.384 & 0.364 \\
          & w/o Audio & 1.660 & 1.607 & 1.590 & 1.540 & 0.570 & 0.559 & 0.538 & 0.481 \\
          & w Incorrect Image** & - & - & 1.423 & - & - & - & 0.381 & - \\
          & w/o Prior Context*** & - & - & 3.428 & - & - & - & 0.478 & - \\
         \hline
         AViD-SP + Meanpool & Base & 1.083 & 1.038 & 0.940 & 0.817 & 0.377 & 0.368 & 0.359 & 0.323 \\
          & w/o Image & 1.051 & 0.980 & 0.911 & 0.826 & 0.386 & 0.377 & 0.362 & 0.330 \\
          & w/o Audio & 0.946 & 0.897 & 0.804 & 0.759 & 0.414 & 0.397 & 0.385 & 0.363 \\
         \hline
    \end{tabular}
    \caption{\label{tab:vg-spice-results}
        RED results on the VG-SPICE test set for our AViD-SP model. AViD-SP was trained with CHiME5 noise augmentation sampled between 0db and 20dB SNR (all CHiME5 noise followed the provided train/eval/test splits). *Given the ground truth utterance transcripts in place of the ASR transcriptions. **Evaluated by offsetting visual features within batch so incorrect image features are paired with the other input components. ***Evaluated with "Empty Context" prior state scene graphs summaries instead of the correct ones. 
    }
\end{table*}

\begin{table}
    \small
    \centering
    \begin{tabular}{l | cc | cc}
        \hline
         Variant & \multicolumn{2}{| c }{TTS} & \multicolumn{2}{| c }{Read} \\
          & H-RED$\downarrow$ & S-RED$\downarrow$ & H-RED$\downarrow$ & S-RED$\downarrow$ \\
         \hline
         GMADS & 0.739 & 0.497 & 0.731 & 0.497 \\
         \hline
         Meanpool & 0.640 & 0.460 & 1.415 & 0.628 \\
         \hline
    \end{tabular}
    \caption{\label{tab:vg-spice-c-results} 
        RED results on the VG-SPICE-C challenge test set for AViD-SP with Single Step (ground truth prior context provided for each step) metrics reported. 
    }
\end{table}

\section{Results} \label{sec:results}

The performance of the AViD-SP model on the VG-SPICE test set, as shown in Table \ref{tab:vg-spice-results}, demonstrates that the baseline AViD-SP achieves S-RED scores just below 0.4, with the meanpooling variant slightly lower, approaching 0.38. This performance suggests a substantial effectiveness (over 60\%) in assimilating desired information into the scene graph. However, the H-RED metrics indicate the introduction of moderate quantities of irrelevant information, particularly in the GMADS version. Given that VG-SPICE scene graphs are often overly sparse, the elevated H-RED values for GMADS may reflect an increased utilization of visual inputs, possibly learning to incorporate non-essential features detected through visual cues. While this interpretation is speculative, some level of elevated H-RED could be reasonable for VG-SPICE in its current state (Section \ref{sec:Qualitative AViD-SP Examples}). 

Under varying SNR conditions, both GMADS and meanpooling configurations of AViD-SP show minimal performance degradation at lower SNRs, indicating resilience to reasonable background noise levels. The use of accurate ASR transcriptions substantially boosts parsing accuracy, emphasizing the benefits of reliable ASR.

Experiments omitting visual inputs or incorporating incorrectly paired visual inputs exhibit minor performance declines. For the meanpooling based AViD-SP a slightly larger, but still quite minor, degradation in metric performance is observed when audio inputs are excluded, with only ASR transcriptions being provided. However, a more significant degradation is observed for the GMADS variant of AViD-SP under these same conditions. This implies that the GMADS multimodal adaptation process has resulted in a model which is more sensitive to the raw audio inputs than when meanpooling is used, which seems to dominantly rely on the natively textual ASR transcriptions. We theorize that the enhanced capability of GMADS to process multimodal inputs may lead to its overall worse results, as it produces a more complex optimization landscape compared with simply collapsing to utilize only the native textual ASR transcripts. Additionally, the absence of prior context markedly increases error rates, underscoring the importance of historical context for accurate SPICE updates.

Table \ref{tab:vg-spice-c-results} presents the performance of AViD-SP on the VG-SPICE-C test set. For TTS audio, the metrics diverge significantly from those of the standard VG-SPICE test set, featuring higher S-RED and lower H-RED scores. The higher density of VG-SPICE-C's scene graphs, which include fewer visually or auditorily supported features that are untracked in reference scene graphs, likely contributes to these lower Hard metric scores. However, this increased density also presents a greater challenge in achieving improved Soft metric scores, as the model must correctly incorporate a substantial amount of information at each update step.

For the GMADS-based AViD-SP, performance metrics on the read audio portion of VG-SPICE-C align closely with those observed in the TTS portion. Conversely, the meanpooling variant shows a substantial performance reduction. This discrepancy suggests that GMADS possesses more robust multimodal processing capabilities, especially in processing out-of-domain real audio distributions. Since both model variants use the same ASR model without parameter tuning, the observed differences indicate that GMADS compensates more effectively for poorer ASR performance.

\section{Conclusion}

In this paper, we introduced Semantic Parsing in Contextual Environments (SPICE), an innovative task designed to enhance artificial agents' contextual understanding by integrating multimodal inputs with prior contexts. Through the development of the VG-SPICE dataset and the Audio-Vision Dialogue Scene Parser (AViD-SP) model, we established a framework for agents to dynamically update their knowledge in response to new information, closely mirroring human communication processes. The VG-SPICE dataset, crafted to challenge agents with the task of visual scene graph construction from spoken conversational exchanges, represents a significant step forward in the field of semantic parsing by incorporating both speech and visual data integration. Meanwhile, the AViD-SP model, equipped with the novel Grouped Multimodal Attention Down Sampler (GMADS), provides a strong initial baseline for VG-SPICE as well as insights into potential methods to improve multimodal information processing and integration.

Our work highlights the importance of developing systems capable of understanding and interacting within complex, multimodal environments. By focusing on the continuous update of contextual states based on new, and multimodal, information, SPICE represents a shift towards more natural and effective human-AI communication. 

\section{Limitations}

While VG-SPICE and AViD-SP are novel approaches, they have several limitations and should be treated as initial attempts toward further SPICE implementations and benchmarks. The main limitation stems from the extensive use of synthetic data augmentation in VG-SPICE's creation. The process involved several steps, including dataset preprocessing with BERT-like POS taggers, crafting update utterances using the Llama 2 70B LLM, and generating synthetic TTS audio. These stages may introduce errors, hallucinations, or overly simple data distributions, potentially misaligning with real-world applications. For example, our models' resilience to background noise may reflect the specific TTS audio distribution, possibly simplifying the ASR model's speech discernment. Additionally, the Visual Genome, our work's foundation, suffers from notable quality issues, such as poor annotations and unreliable synthetic object segmentation, which, despite efforts to mitigate, remain challenges in VG-SPICE. While the included VG-SPICE-C test subset attempts to improve these limitations, and indeed the hard versions of are metrics are significantly improved on the manually cleaned samples of this subset, they are still comprised of intentionally crafted utterances with read audio, which may not transfer to real-world applications and natural spoken audio. Further, this work only includes analysis of the VG-SPICE-C challenge subset in the simple Single Step task and does not evaluate in end-to-end sequence-based analysis. 

The various version of AViD-SP we introduce also provides indications of further development for efficient multimodal adaptation methodologies. While the version utilizing GMADS generally failed to outperform the results of the traditional meanpooling version the GMADS method also provided a stronger indication of cross-modality feature utilization, whereas integration of simplistically downsampled multimodal features alongside native textual features appears to cause strong underutilization and feature collapse for the multimodal features. This is further supported by the poor performance achieved by the meanpooling version of AViD-SP, relative to the GMADS version, on real human recorded audio, indicating the meanpooling version adapts much worse to out-of-domain multimodal inputs. We suggest future work to continue investigating methods similar to GMADS to further realize their theoretical benefits. 

Moreover, VG-SPICE, while pioneering in SPICE tasks, is only a start, limited to audio and images, with a basic language for knowledge graph updates. Future research should address these limitations by incorporating more realistic inputs, like video, 3D environments, and paralinguistic cues, and by exploring dynamic tasks beyond simple scene graph updates. Environments like Matterport3D \citep{Matterport3D} or Habitat 3.0 \citep{puig2023habitat} offer promising avenues for embodied SPICE research. Expanding SPICE to include secondary tasks that rely on an agent's contextual understanding can also enhance its utility, such as aiding in medical image annotation with co-dialogue.

\nocite{}

\bibliography{acl_latex}

\appendix

\section{Acknowledgements}
This material is based upon work supported by the National Science Foundation under Grant No. 2238605

\section{Additional AViD-SP Results} \label{sec:Additional AViD-SP Results}
We report the Graph Edit Distance (GED) results for AViD-SP, and the tested baselines, here.

\begin{table*}
    \small
    \centering
    \begin{tabular}{lr | cccc | cccc}
        \hline
         \multicolumn{2}{l}{Model Type} & \multicolumn{4}{| c |}{H-GED$\downarrow$} & \multicolumn{4}{| c }{S-GED$\downarrow$} \\
         \multicolumn{2}{l |}{} & 0dB & 2dB & 20dB & Gold* & 0dB & 2dB & 20dB & Gold* \\
         \hline
         AViD-SP + GMADS & Base & 2.010 & 1.921 & 1.811 & 1.621 & 0.924 & 0.889 & 0.862 & 0.778 \\
          & w/o Image & 2.044 & 1.973 & 1.816 & 1.642 & 0.944 & 0.923 & 0.878 & 0.791 \\
          & w/o Audio & 2.168 & 2.101 & 2.071 & 1.863 & 1.209 & 1.186 & 1.158 & 1.004 \\
          & w Incorrect Image** & - & - & 1.806 & - & - & - & 0.861 & - \\
          & w/o Prior Context*** & - & - & 4.656 & - & - & - & 0.909 & - \\
         \hline
         AViD-SP + Meanpool & Base & 1.739 & 1.617 & 1.514 & 1.295 & 0.935 & 0.889 & 0.859 & 0.759 \\
          & w/o Image & 1.732 & 1.599 & 1.514 & 1.285 & 0.939 & 0.910 & 0.872 & 0.759 \\
          & w/o Audio & 1.622 & 1.560 & 1.428 & 1.244 & 1.002 & 0.964 & 0.909 & 0.815 \\
          & w Incorrect Image** & - & - & 1.517 & - & - & - & 0.857 & - \\
          & w/o Prior Context*** & - & - & 4.778 & - & - & - & 0.905 & - \\
         \hline
    \end{tabular}
    \caption{\label{tab:vg-spice-ged-results}
        GED results on the VG-SPICE test set for our AViD-SP model. AViD-SP was trained with CHiME5 noise augmentation sampled between 0db and 20dB SNR (all CHiME5 noise followed the provided train/eval/test splits). *Given the ground truth utterance transcripts in place of the ASR transcriptions. **Evaluated by offsetting visual features within batch so incorrect image features are paired with the other input components. ***Evaluated with "Empty Context" prior state scene graphs summaries instead of the correct ones. 
    }
\end{table*}

\section{Qualitative AViD-SP Examples} \label{sec:Qualitative AViD-SP Examples}

\begin{figure*}
\centering
\includegraphics[width=1.0\linewidth]{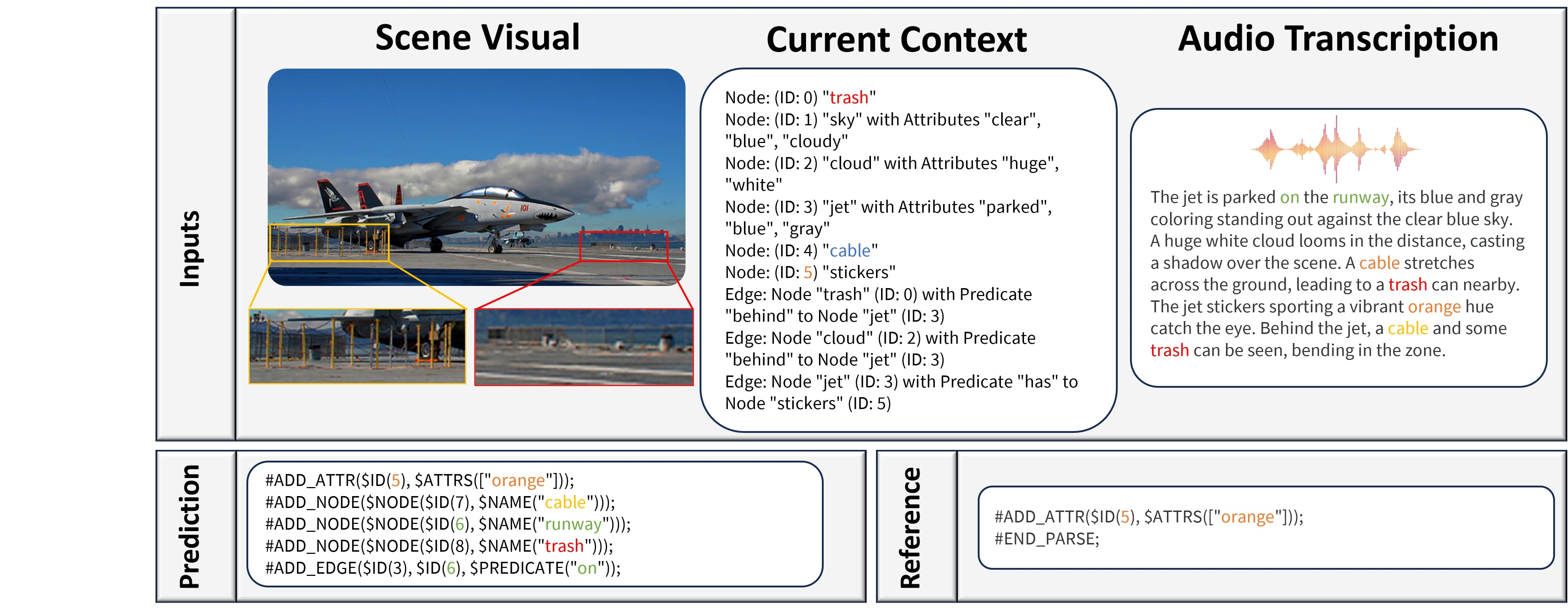}
\caption{Sample generation output with corresponding inputs from AViD-SP. Scored a Soft-RED of 0.0 and Hard-RED of 6.727. Significant features highlighted in colors. Qualitative evaluation reveals that the majority of extraneous additions were either supported by the Audio Transcription, the scene image, or both. }
\label{fig:qual-ex}
\end{figure*}

We include an example of a typical AViD-SP generation in Figure \ref{fig:qual-ex}, with metric scores approximately at the average obtained across the full testing set. In this example it is evident that all of the ground truth reference information was successfully added to the updated scene graph, leading to the Soft-RED score of 0.0. However, considerable extraneous information is also observed to have been added. In Figure \ref{fig:qual-ex} three additional Nodes are added, with two of them being duplicates of ones that already exist in the scene graph, along with one Edge. 

However, considering the Transcription and Visual Scene for the illustrated sample reveals that these features, while not included in the reference, likely are logically reasonable for the agent to include. For the additional Node of ``runway'' the motivation is obvious. Not only is the runway and its corresponding edge relationship mentioned by the LLM, but a runway is even present in the scene visual. Similar conditions apply to the two duplicate nodes added. While those nodes already exist, they are mentioned in the Audio Transcription at two distinct times. Inspection of the highlighted and blown-up parts of the image also reveals that there are in fact duplicates of these entities in the scene, making their addition to the updated context reasonable. 

This is not to say all extraneous additions should be treated as correct since many should not. However, it does illustrate a key area to seek further improvement in the VG-SPICE dataset and why, for this work, we focus more on the ``soft'' capability to add all known good information tot he graph. 

\section{Visual Genome Preprocessing} \label{sec: Visual Genome Preprocessing} 

The Visual Genome serves as a strong basis for VG-SPICE but has quality issues such as inconsistent naming for Nodes, Attributes, and Predicates, duplicate Nodes, and unnecessary Nodes (e.g., \textbf{<man, has, head>}). Prior solutions for Scene Graph Generation (SGG) tasks \citep{liang2019vrrvg, zhang2019largescale, xu2017scene, maëlic2023finegrained} curated versions by limiting predicates and node names, reducing predicates from 27k to 50 and node names from 53k to 150. While the Visual Genome contains a substantial portion of single-sample terms, typically of lower quality, such restrictions can oversimplify and yield smaller, less representative scene graphs.

Our approach refines the Visual Genome by:

\paragraph{Standardization and Correction:} We applied rule-based systems with Sentence Transformer Part of Speech taggers \footnote{Using "all-mpnet-base-v2" from Python Sentence Transformers} to fix inconsistencies and improve scene graph density by retaining rare Node names (e.g., "red table", identifying "red" as an attribute). We removed low-quality attributes and predicates by limiting them to specific parts of speech conditions, such as removing proper and common nouns from attributes/edges. Furthermore, we imposed several straightforward constraints to refine the scene graph structure. These included setting limits on the word counts for individual scene graph elements and consolidating attributes when redundancy was detected within a specific node, for instance, merging "reddish" and "red" when both attributes described the same entity.
\paragraph{Duplicate Node Elimination:} We added a post-standardization phase to remove duplicate nodes. Unlike earlier methods \citep{maëlic2023finegrained} relying solely on a high Intersection over Union (IoU) threshold for exact node matches, we included a semantic similarity check from the contextualized embeddings from the same Sentence Transformer utilized in the Standardization and Correction phase. This allows for the detection of duplicate Nodes with significant name similarities and IoUs. With a preference for visually supported scene graphs over the potential exclusion of some valid Nodes, we set a lower IoU threshold (0.5, compared to prior works' 0.9) and a semantic similarity threshold of 0.7.
\paragraph{Term Frequency Analysis:} Next, we manually curated terms in the filtered dataset to establish a relevant set for the SPICE task, excluding single-occurrence terms for their low quality, and filtered scene graphs based on this list.
\paragraph{Scene Graph Size Restriction:} Finally, we filtered out small graphs to ensure a diverse set for VG-SPICE, excluding graphs with fewer than four Nodes or Edges and applying dynamically increased threshold for graphs with duplicate nodes.

These methods enhanced the Visual Genome's graphs, yielding a dataset with improved quality and annotation density, as illustrated in Table \ref{tab:vg-clean}.

\section{Training and Inference Hyperparameters} \label{sec: Training and Inference Hyperparameters}
The training regimen for AViD-SP spans two epochs across the dataset, using a combined batch size of 72 on six Nvidia L40 GPUs. An initial learning rate of $5 \times 10^{-5}$ is applied, followed by exponential decay. We employ cross-entropy loss for the prediction of target semantic parses, introducing loss masking for padding and for the prompt that combines prior context with multimodal inputs. We utilize loss factors of $\alpha=1.0$, $\beta=0.1$, and $\gamma=0.1$.

Inference leverages a greedy decoding strategy with a max generation length of 160 tokens and otherwise default generation parameters for LLAMA 2 7B. 

\section{Contextual State Representation} \label{sec: Contextual State Representation}

SPICE formulates the prior context to be utilized by the agent as a structured knowledge graph. However, top-performing semantic parsing generation models, such as those best on the Llama architecture as used in this work, are decoder-only models that can accept inputs from linear text sequences only. This requires utilizing either a compatible knowledge graph encoder which can embed and project the knowledge graph representation for use by the semantic parse generation model, or representing the knowledge graph in the form of a textually formatted prompt. For AViD-SP developed in this work, we utilized the second, with the format of the textually prompted representation of the prior context shown in Figure \ref{fig:vg-spice-examples}.

When generating the context representations all existing Nodes are assigned Node IDs, and semantic parses are expected to operate in reference to these Node IDs (Section \ref{sec: Formal Language Definition}). We provide Nodes and Attributes first, followed by any Edges. The ordering of all information is sorted by Node ID in ascending order. In practice, all Node IDs are randomly assigned for each training iteration to diversity training inputs. 

\section{Formal Language Definition} \label{sec: Formal Language Definition}
The formal language we used in the semantic parses $P_{i}$ and the corresponding execution function $e$ contained the following executable function, which together could deterministically update the scene graph prior context $C_{i}$ to the next context state $C_{i+1}$. Since VG-SPICE only represents the conversational construction of scene graphs, and not deletion or alterations, our formal language is comprised of three distinct operations: 1) \textit{\#ADD\_NODE} accepting a new Node ID, name, and optionally a set of attributes to add along with it, 2) \textit{\#ADD\_ATTR} accepting an existing Node ID as well as a set of attributes to be added to the specified node, and 3) \textit{\#ADD\_EDGE} accepting a source and target pair of existing node IDs along with the predicate to be assigned between them. Our formal language always generates reference semantic parses with new attributes added first, followed by new Nodes (and assigned attributes), and lastly new edges. However, when evaluating our model outputs the execution function $e$ can accept these commands in any order, so long as the referenced node IDs already have been added. 

\section{Licensing}

Our paper utilized the Visual Genome dataset which is listed under a Creative Commons license. All other tools utilized are available from either Pythons Spacy or Huggingface and are available for academic use. To the best of our knowledge, all artifacts utilized are aligned with their intended use cases. 

\section{AI Assistance}

A minor portion of code development was done with the assistance of ChatGPT. All research ideas and writing are of the author's original creation. Grammarly was utilized for writing assistance.

\end{document}